\documentclass[runningheads]{llncs}
\usepackage{graphicx}
\usepackage{graphbox} %
\usepackage{array,etoolbox}
\preto\tabular{\setcounter{magicrownumbers}{0}}

\usepackage{pifont}%
\newcommand{\cmark}{\ding{51}}%
\newcommand{\xmark}{\ding{55}}%
\newcommand{\vs}{\emph{vs }}
\usepackage{caption}
\usepackage{color}

\usepackage[numbers,sort]{natbib} %

\begin{document}

\title{GhostVLAD for set-based face recognition} 
\titlerunning{GhostVLAD for set-based face recognition} %

\author{Yujie Zhong\inst{1} \and
Relja Arandjelovi\'c\inst{2}\and
Andrew Zisserman\inst{1,2}}

\authorrunning{Y. Zhong, R. Arandjelovi\'c and A. Zisserman} 

\institute{\hspace{-2mm}VGG, Department of Engineering Science, University of Oxford, UK\\
\email{\{yujie,az\}@robots.ox.ac.uk} \and 
\hspace{-2mm}DeepMind\\
\email{relja@google.com}}

\def\eg{\emph{e.g. }}
\def\Eg{\emph{E.g. }}
\def\etal{\emph{et al. }}
\def\ie{\emph{i.e. }}
\def\etc{\emph{etc. }}

\renewcommand{\paragraph}[1]{\medskip\noindent{\emph{#1}}}

\maketitle

\begin{abstract}

The objective of this paper is to learn a 
compact representation of image sets for
template-based  face recognition.
We make the following 
contributions:
first, we propose a network architecture which 
aggregates and  embeds the face descriptors 
produced by deep convolutional neural networks 
into a compact fixed-length representation.
This compact representation requires minimal memory 
storage and enables  efficient similarity 
computation. 
Second, we propose a novel 
GhostVLAD layer that includes {\em ghost clusters}, that do not contribute to the aggregation.
We show that a quality
weighting on the input faces  emerges automatically
such that informative images 
contribute more than those with low  quality, and 
that  the ghost clusters enhance  the network's  ability to 
deal with poor quality images. 
Third, we explore how input feature dimension,
number of clusters and 
different training techniques affect the recognition
performance. Given this analysis, we train
a network that far exceeds the state-of-the-art
on the IJB-B face recognition dataset. This  is
currently one of the most challenging public benchmarks, and we surpass the state-of-the-art on both
the identification and verification protocols.

\end{abstract}

\section{Introduction}
\label{sec:intro}
While most research on face recognition 
has  focused on recognition from a single-image, 
 {\em template} based  face recognition, where a set of faces of the same subject is available,
is now gaining attention.
In the  unconstrained scenario considered here, this  can be a challenging task as
face images may have 
various poses,  expression, illumination, and may also be of quite varying quality.

A straightforward method to tackle multiple images per 
subject is to store per-image descriptors extracted 
from each face image (or frame in a video), and 
compare every  pair of images between sets at query 
time~\cite{Taigman14,Schroff15}. 
However, this type of approach can be 
memory-consuming and prohibitively 
slow, especially for searching tasks in large-scale datasets.
Therefore, an aggregation method that can 
produce a compact template representation is desired. 
Furthermore, this representation should 
support efficient computation of similarity 
and require minimal memory storage. 

More importantly, the representation 
obtained from image sets should be 
discriminative 
\ie template descriptors of the same subject 
should be close to each other in the descriptor space, 
whereas those of different subjects should be far apart.
Although common aggregation strategies, such as 
average pooling and max-pooling, are able to
aggregate face descriptors to
produce a compact template 
representation~\cite{Parkhi15,Chen15,Cao18} 
and currently achieves the state-of-the-art results~\cite{Cao18},
we seek a better solution in this paper.

As revealed by~\cite{Jegou14}, image retrieval encoding 
methods like Fisher Vector encoding 
and T-embedding increase the separation between 
descriptors extracted from related 
and unrelated image patches. 
We therefore expect a similar encoding to be beneficial 
for  face recognition, including both verification and 
identification tasks. This insight inspires us 
to include a similar encoding, NetVLAD~\cite{Arandjelovic16},  in the design of our 
network.

In this paper, we propose a convolutional neural network (Fig.~\ref{fig:cnn})
that satisfies all the desired properties mentioned above: 
it can take any number of input faces and 
produce a compact fixed-length descriptor to represent 
the image set. Moreover, this network
embeds face descriptors such that the resultant template-descriptors 
are  more discriminative than the original descriptors.
The 
representation is efficient in both 
memory and query speed aspects, \ie  it only stores 
one compact descriptor per template, regardless of the number 
of face images in a template, and the similarity 
between two templates is simply measured as the 
scalar product (\ie cosine similarity) 
of two template descriptors. 

However, one of the key problems  in unconstrained real-world 
situations is that some faces in a template may be of low quality -- for example,
low resolution, or blurred, or partially occluded.
These low-quality images are distractors 
and are likely to hurt the performance of the 
face recognition system if given equal weight as the other (good quality) faces. Therefore, a sophisticated 
network should be able to reduce the impact 
of such distracting images and focus on the informative 
ones. 

To this end, we extend the NetVLAD architecture to include 
{\em  ghost clusters}. These are clusters that face descriptors can be soft assigned to, but are excluded from 
the aggregation. They provide a mechanism for the network to handle low quality faces, by mainly assigning
them to the ghost clusters.
Interestingly, although we do not explicitly learn 
any importance weightings between faces in each template, 
such property emerges automatically from our network. 
Specifically,
low quality  faces generally contribute less to the 
final template representation than the high-quality ones.

The networks are trained in an 
end-to-end fashion with only identity-level labels.
They outperform 
state-of-the-art methods by a large margin
on the public IJB-A~\cite{Klare15} 
and IJB-B~\cite{Whitelam17} face recognition benchmarks.
These datasets are currently the 
most challenging  in the community, and
we evaluate on these  in this paper.

This paper is organized as following: Sec.~\ref{sec:review}
reviews some related work on face recognition based on 
image sets or videos; the proposed network  
and implementation details are introduced
in Sec.~\ref{sec:cnn}, followed by
experimental results reported
in Sec.~\ref{sec:exp}. Finally a conclusion is drawn 
in Sec.~\ref{sec:conclusion}.

\section{Related work}
\label{sec:review}
Early face recognition approaches which make use of sets of face examples
(extracted from different images or video frames)
aim to represent
image sets as 
manifolds~\cite{Lee03,Arandjelovic06,Kim07,Wang08,Turaga11,Yang13,Huang15},
convex hulls~\cite{Cevikalp10},
Gaussian Mixture Models~\cite{Wang15a},
or set covariance matrices~\cite{Wang17a}, 
and measure the dissimilarity between image sets 
as distance
between these spaces.

Later methods represent face sets
more efficiently using a single fixed-length descriptor.
For example, \cite{Parkhi14} aggregates local descriptors
(RootSIFT \cite{Arandjelovic12}) extracted 
from face crops using
Fisher Vector~\cite{Perronnin10} (FV) 
encoding to obtain a single 
descriptor per face track. Since the success of 
deep learning in image-based face 
recognition~\cite{Taigman14,Parkhi15,Schroff15,Sun15,Liu17b,Zheng18,Cao18a}, 
simple strategies for face descriptor aggregation prevailed,
such as average- and 
max-pooling~\cite{Parkhi15,Chen15}.
However, none of these strategies are trained end-to-end
for face recognition as typically only the face descriptors
are learnt, while aggregation is performed post hoc.

A few methods go beyond simple pooling by computing a weighted
average of face descriptors based on some measure of
per-face example importance.
For example, \cite{Goswami14} train a module to predict human
judgement on how memorable a face is, and use this memorability score 
as the weight.
In~\cite{Yang17},
an attention mechanism is used to compute face example weights,
so that the contribution of low quality images
to the final set representation is down-weighted.
However, these methods rely on pretrained face descriptors
and do not learn them
jointly with the weighting functions,
unlike our method where the entire system is trained end-to-end
for face recognition.

Two other recent papers are quite related in that 
they explicitly take account of image quality:
\cite{Hassner16} first bins 
face images of similar quality and pose before aggregation; whilst~\cite{Liu17} introduces a fully end-to-end trainable
method which automatically learns to down-weight low quality images. As will be seen in the sequel, we achieve similar 
functionality implicitly due to the network architecture, and also exceed the performance of both these methods
(see Sec.~\ref{sec:exp:sota}).
As an interesting yet different method 
which can also filter low-quality images, 
\cite{Rao17} learns to  aggregate  the raw face images and then computes a 
descriptor.

Our aggregation approach is inspired by the image retrieval literature
on aggregating local descriptors~\cite{Jegou14,Arandjelovic16}.
Namely, J\'egou and Zisserman \cite{Jegou14} find that,
compared to simple average-pooling,
Fisher Vector encoding and T-embedding increase
the contrast between the similarity scores of 
matching and mismatching local descriptors.
Motivated by this fact, we make use of a trainable aggregation layer,
NetVLAD~\cite{Arandjelovic16}, and improve it for the face recognition task.

\section{Set-based face recognition}
\label{sec:cnn}
We aim to learn a compact representation of a face.
Namely, we train a network which digests a set of example face images of
a person, and produces a fixed-length template representation
useful for face recognition.
The network should satisfy the following properties:

(1) Take any number of images as input, and output a 
fixed-length template descriptor to represent the 
input image set.
(2) The output template descriptor should be compact
(\ie low-dimensional) in order to require little memory
and facilitate fast template comparisons.
(3) The output template descriptor should be 
discriminative, such that the similarity of
templates of the same subject is much larger than that
of different subjects.

We propose a convolutional neural 
network that fulfils all three objectives.
(1) is achieved by aggregating face descriptors
using a modified NetVLAD~\cite{Arandjelovic16} layer, GhostVLAD.
Compact template descriptors (2) are produced by a
trained layer which performs dimensionality reduction.
Discriminative representations (3) emerge because
the entire network is trained end-to-end for face recognition,
and since our GhostVLAD layer is able to
down-weight the contribution of low-quality images,
which is important for good
performance~\cite{Goswami14,Yang17,Hassner16}.

The network architecture and the new GhostVLAD layer
are described
in Sec.~\ref{sec:arch} and Sec.~\ref{sec:ghost}, respectively,
followed by 
the network training procedure (Sec.~\ref{sec:train})
and implementation 
details (Sec.~\ref{sec:imp}).

\subsection{Network architecture}
\label{sec:arch}
\begin{figure}[t]
   \begin{center}
         \includegraphics[width=1\columnwidth]{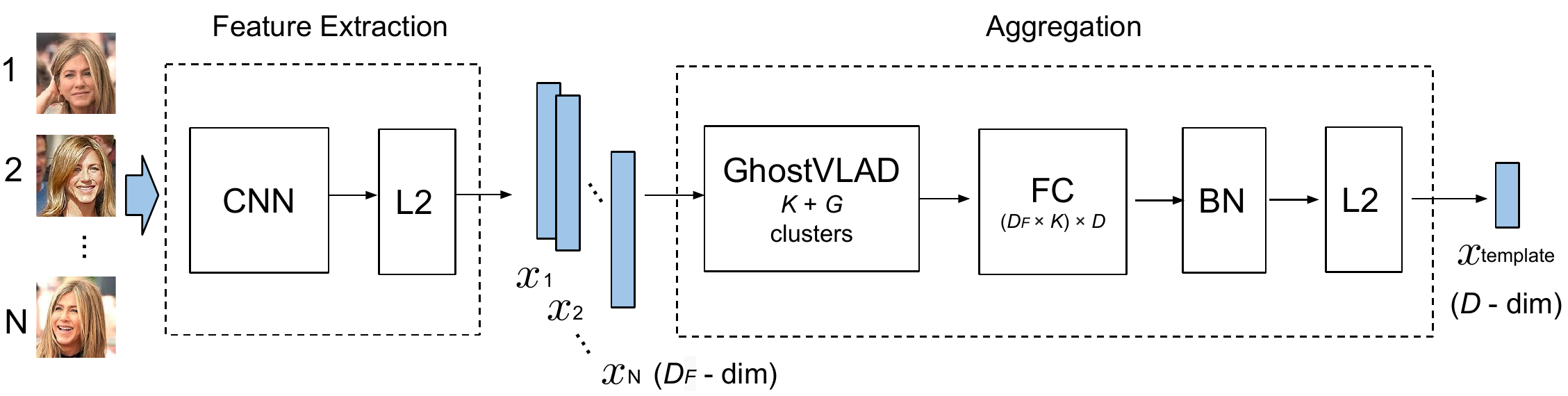}
   \end{center}
   \caption{\textbf{Network architecture.}
Input images in each template are first 
passed through a convolutional 
neural network (\eg ResNet-50 or SENet-50 with an additional
FC layer and L2-normalization) to produce 
a face descriptor per image.
The descriptors are aggregated into a single fixed-length
vector using the GhostVLAD layer.
The final $D$-dimensional template descriptor
is obtained by reducing dimensionality using a fully-connected layer,
followed by batch normalization (BN) and
L2-normalization.
   }
    \label{fig:cnn}
\end{figure}

As shown in Fig.~\ref{fig:cnn},
the network consists of two parts:
feature extraction, which computes a 
face descriptor for each input face image, and 
aggregation, which aggregates all face descriptors
into a single compact template representation of
the input image set.

\paragraph{Feature extraction.}
A neural network is used to extract
a face descriptor for each input face image.
Any network can be used in our learning framework,
but in this paper we opt for
ResNet-50~\cite{He16} or SENet-50~\cite{Hu18}.
Both networks are cropped 
after the global average pooling layer,
and an extra FC layer is added to reduce the 
output dimension to $D_F$.
We typically pick $D_F$ to be low-dimensional (\eg 128 or 256),
and do not see
a significant drop in face recognition performance
compared to using the original 2048-D descriptors.
Finally, the individual face descriptors are L2 normalized.

\paragraph{Aggregation.}
The second part uses GhostVLAD (Sec.~\ref{sec:ghost})
to aggregate multiple face descriptors into a single
$D_F \times K$ vector (where $K$ is a parameter of the method).
To keep computational and memory requirements low,
dimensionality reduction is performed via an FC layer,
where we pick the output dimensionality $D$ to be 128.
The compact $D$-dimensional
descriptor is then passed to a batch-normalization
layer~\cite{Ioffe15} and L2-normalized 
to form the final template 
representation $x_{template}$.

\subsection{GhostVLAD: NetVLAD with ghost clusters} \label{sec:ghost}
The key component of the aggregation block is our \emph{GhostVLAD}
trainable aggregation layer, which given $N$ $D_F$-dimensional face descriptors computes a single $D_F \times K$ dimensional output.
It is based on the NetVLAD~\cite{Arandjelovic16} layer
which implements an encoding similar to 
VLAD encoding~\cite{Jegou10}, while being 
differentiable and thus fully-trainable.
NetVLAD  has been shown to outperform average and max pooling for the same
vector dimensionality, which makes
it perfectly suited for our task.
Here we provide a brief overview of NetVLAD
(for full details please refer to~\cite{Arandjelovic16}),
followed by our improvement, GhostVLAD.

\paragraph{NetVLAD.}
For $N$ $D_F$-dimensional input descriptors $\{x_i\}$ and
a chosen number of clusters $K$, NetVLAD pooling produces a single $D_F \times K$
vector $V$ (for convenience written as a $D_F \times K$ matrix) according
to the following equation:
\begin{equation}
V(j,k) = \sum_{i=1}^{N} \frac{e^{a_k^Tx_i + b_k}}{\sum_{k'=1}^{K}{e^{a_{k'}^Tx_i + b_{k'}}}} (x_i(j) - c_k(j))
\label{eq:netvlad}
\end{equation}
where $\{a_k\}$, $\{b_k\}$ and $\{c_k\}$ are trainable parameters, with $k \in [1, 2, \dots, K]$.
The first term corresponds to the soft-assignment weight
of the input vector $x_i$ for cluster $k$,
while the second term computes the residual between the vector and
the cluster centre.
The final output is obtained by performing L2 normalization.

\begin{figure}[t]
   \begin{center}
         \includegraphics[width=0.99\columnwidth]{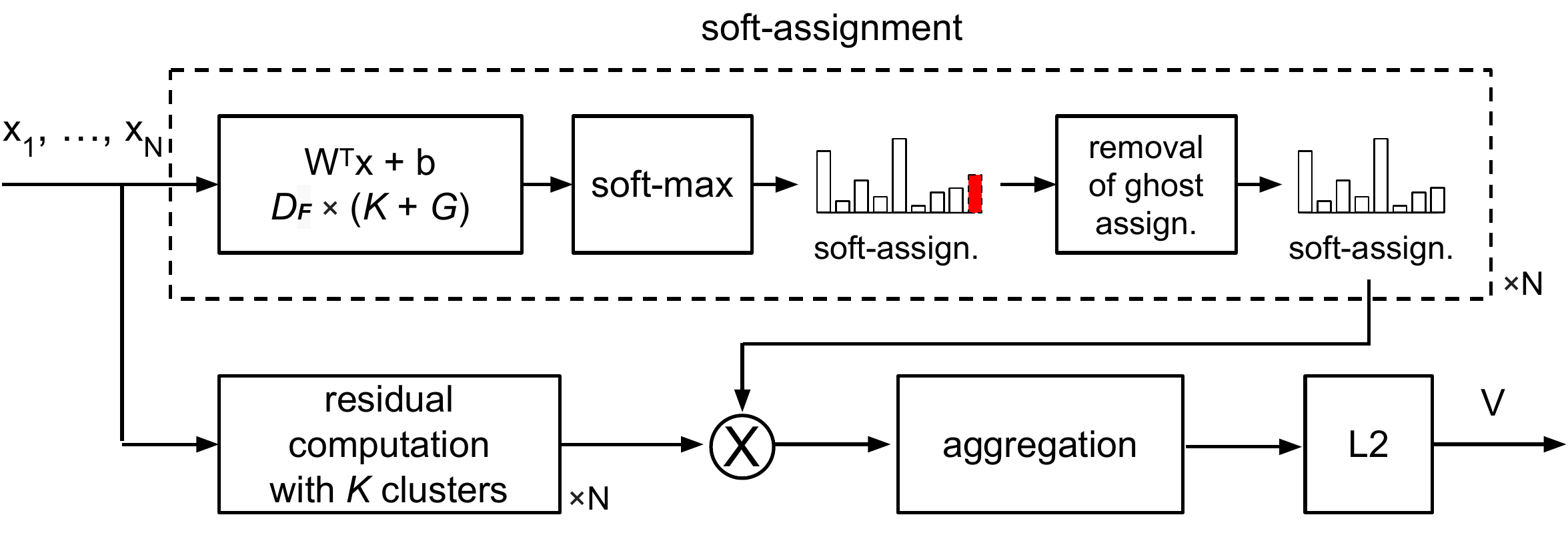}
   \end{center}
   \caption{\textbf{GhostVLAD.}
For each input descriptor, NetVLAD performs soft-assignment
into $K$ cluster centres, computed as a linear transformation
followed by a soft-max.
It then, for each cluster centre, aggregates all residuals 
between input descriptors and the cluster centre,
weighted with the soft-assignment values.
The final vector is produced as a concatenation of the per-cluster
aggregated residuals;
for more details see eq.~\ref{eq:netvlad} and~\cite{Arandjelovic16}.
We introduce $G$ ``ghost'' clusters in the soft-assignment stage,
where the ``ghost'' assignment weight is illustrated
with a dotted red bar (here we show only $G=1$ ghost cluster).
The ghost assignments are then eliminated and residual aggregation
proceeds as with NetVLAD.
This mechanism enables the network to assign uninformative descriptors
to ghost clusters thus decreasing their soft-assignment weights
for non-ghost clusters, and therefore reducing their contribution
to the final template representation.
   }
    \label{fig:ghost}
\end{figure}

\paragraph{GhostVLAD.}
We extend NetVLAD with ``ghost'' clusters to form \emph{GhostVLAD},
as shown in Fig.~\ref{fig:ghost}.
Namely, we add further $G$
``ghost'' clusters which contribute to the soft assignments in the
same manner as the original $K$ clusters,
but residuals between input vectors and the ghost cluster centres
are ignored and do not contribute to the final output.
In other words, the summation in the denominator of
eq.~\ref{eq:netvlad} instead of to $K$ goes to $K+G$,
while the output is still $D_F \times K$ dimensional;
this means $\{a_k\}$ and $\{b_k\}$ have $K+G$ elements each,
while $\{c_k\}$ still has $K$.
Another view is that we are computing NetVLAD with $K+G$ clusters,
followed by removing the elements that correspond to the $G$
ghost clusters.
Note that GhostVLAD is a generalization of NetVLAD as with $G=0$
the two are equivalent.
As with NetVLAD, GhostVLAD can be implemented efficiently using
standard convolutional neural network building blocks,
\eg the soft-assignment can be done by stacking input descriptors
and applying a convolution operation, followed by a convolutional soft-max;
for details see~\cite{Arandjelovic16}.

The intuition behind the incorporation of ghost clusters is to
make it easier for the network to adjust
the contribution of each face example to the template representation
by assigning examples to be ignored to the ghost clusters.
For example, in an ideal case, a highly blurry face image
would be strongly assigned to a ghost cluster,
making the assignment weights to non-ghost clusters close to zero,
thus causing its contribution to the template representation
to be negligible;
in Sec.~\ref{sec:visualize_ghost} we qualitatively
validate this intuition.
However, note that we do not explicitly force low-quality images to
get assigned to ghost clusters, but instead let the network discover
the optimal behaviour through end-to-end training for face recognition.

\subsection{Network training}
\label{sec:train}
In this section we describe how to train the network for
face recognition, but note that GhostVLAD is a general layer which can
also be used for other tasks.

\paragraph{Training loss.}
Just for training purposes, we append the network with a fully-connected
``classification'' layer of size $D \times T$, where $D$ is the size
of the template representation and $T$ is the number of identities
available in the training set.
We use the one-versus-all logistic regression loss as empirically
we found that it converges faster and
outperforms cross-entropy loss.
The classification layer is discarded after training and the trained network
is used to extract a single fixed-length template representation
for the input face images.

\paragraph{Training with degraded images.}
For unconstrained face recognition, it is important to be able to handle
images of varying quality that typically occur in the wild.
The motivation behind our network architecture, namely the GhostVLAD layer,
is to enable it to down-weight the influence
of these images on the template representation.
However, since our training dataset only contains good quality images,
it is necessary to perform data augmentation in the form of image degradation,
such as blurring or compression (see Sec.~\ref{sec:imp} for details),
in order to more closely match the varying image quality encountered
at test time.

\subsection{Implementation details}
\label{sec:imp}
This section discusses full details of the 
training process, including training data,
data augmentation,
network initialization, \etc

\paragraph{Training data.} 
We use face images from the training set of the
VGGFace2 dataset~\cite{Cao18} to train the network.
It consists of around 3 million images, covering 
8631 identities. For 
each identity, there are on average 360 face 
images across different ages and poses.
To perform set-based training,
we form image sets on-the-fly by repeatedly sampling
a fixed number of images belonging to the same identity.

\paragraph{Data augmentation.}
Training images are resized such that the smallest 
dimension is 256 and random crops of size 
$224 \times 224$ are used as inputs to the 
network. Further augmentations include random
horizontal flipping and a random rotation of 
no greater than 10 degrees.

We adopt four methods to degrade images for training:
isotropic blur, motion blur, decreased resolution and JPEG compression.
Each degradation method has a 
probability 
of 0.1 to be applied to a training image,
where, to prevent over-degradation, a maximum of
two transformations per image is allowed.
Isotropic blur is implemented using a Gaussian filter
where the standard deviation is uniformly sampled between 6 and 16.
For motion blur, the angle of motion is uniformly sampled
between 0 and 359 degrees, and the motion length is fixed to 11. 
Resolution decrease is simulated by downscaling the image by a factor of 10
and scaling it back up to the original size.
Finally, we add JPEG compression artefacts by randomly compressing
the images to one of three compression ratios: 0.01, 0.05 and 0.09.

\paragraph{Training procedure.}
The network can be trained 
end-to-end in one go, but, to make the training faster and
more stable, we divide it
into three stages; all stages only use the VGGFace2~\cite{Cao18}
dataset.
In the first two stages, parts of the network are trained
for single-image face classification
(\ie the input image set consists of a single image),
and image degradation is not performed.
Firstly, the feature extractor network
is pre-trained for single-image face classification
by temporarily (just for this stage)
appending a classification FC layer on top of it,
and training the network with the cross-entropy loss.
Secondly, we train the whole network
end-to-end for single-image classification
with one-versus-all logistic regression loss,
but exclude the ghost clusters from GhostVLAD
because training images are not degraded in this stage.
Finally, we add ghost clusters and enable image degradation,
and train the whole network using image
sets with one-versus-all logistic regression loss.

\paragraph{Parameter initialization.}
The non-ghost clusters of GhostVLAD are initialized
as in NetVLAD~\cite{Arandjelovic16} by clustering
its input features with k-means into $K$ clusters,
where only non-degraded images are used.
The $G$ ghost clusters are initialized similarly, but using
degraded images for the clustering;
note that for $G=1$ (a setting we often use)
k-means simplifies to computing the mean over the features.
The FC following GhostVLAD which performs dimensionality reduction
is then initialized using 
the PCA transformation matrix computed on the 
GhostVLAD output features.

\paragraph{Training details.}
The network is trained using stochastic gradient 
descend with momentum,
implemented in MatConvNet~\cite{Vedaldi15}.
The mini-batch consists of 84 face images,
\ie if we train with image sets of size two,
a batch contains 42 image sets, one per identity.
When  
one-versus-all logistic regression loss is used,
for each image set, we update the network weights 
based on the positive class and only 20 negative 
classes (instead of 8631) that 
obtain the highest classification scores.
The initial learning rate of 0.0001 is used for all
parameters apart from GhostVLAD's assignment parameters
and the classification FC weights,
for which we use 0.1 and 1, respectively.
The learning rates are divided by 10 when validation error stagnates,
while weight decay and momentum are fixed to 0.0005 and 0.9,
respectively.

\section{Experiments}
\label{sec:exp}
In this section, we describe the experimental setup,
investigate the impact of our design choices,
and compare results with the state-of-the-art.

\subsection{Benchmark datasets and evaluation protocol}
Standard and most challenging public face recognition datasets
IJB-A~\cite{Klare15} and IJB-B~\cite{Whitelam17}
are used for evaluation.
In contrast to single-image based face datasets such as
\cite{Parkhi15,Guo16,Kemelmacher16,Cao18},
IJB-A and IJB-B are intended for template-based face recognition,
which is exactly the task we consider in this work.
The IJB-A dataset contains 5,712 images 
and 2,085 videos, covering 500 subjects; thus the
average number of images and videos per subject
are 11.4 and 4.2 videos, respectively.
The IJB-B dataset is an extension of IJB-A
with a total of 11,754 images and
7,011 videos
from 1,845 subjects, as well as 
10,044 non-face images.
There is no overlap between subjects in VGGFace2, which we use
for training, and the test datasets.
Faces are detected from images and all video frames
using MTCNN~\cite{Zhang16},
the face crops are then resized such that the smallest 
dimension is $224$ and the central $224 \times 224$ crop is
used as the face example.

\paragraph{Evaluation protocol.}
We follow the standard benchmark procedure for
IJB-A and IJB-B, and evaluate on
``1:1 face verification'' and ``1:N face identification''.

The goal of \emph{1:1 verification} is to make a decision
whether two templates belong to the same person,
done by thresholding the similarity between the templates.
Verification performance is assessed via the
receiver operating characteristic (ROC) curve,
\ie by measuring the trade-off between the
true accept rates (TAR) \vs false accept 
rates (FAR).

For \emph{1:N identification},
templates from the probe set are used to rank
all templates in a given gallery. 
The performance is 
measured using the true positive identification 
rate (TPIR) \vs false positive identification 
rate (FPIR) (\ie the decision error trade-off (DET) curve)
and \vs Rank-N (\ie the cumulative match characteristic 
(CMC) curve).

Evaluation protocols are the same for both 
benchmark datasets, apart from the fact that
IJB-A defines 10 test splits,
while IJB-B only has one split for verification
and two galleries for identification.
For IJB-A and for IJB-B identification, we report,
as per standard, the mean and standard deviation of
the performance measures.

\subsection{Networks, deployment and baselines}

\paragraph{Our networks.}
As explained earlier in Sec.~\ref{sec:arch},
we use two different
architectures as backbone feature extractors:
ResNet-50~\cite{He16} 
and SENet-50~\cite{Hu18}.
They are cropped after global average-pooling
which produces a $D_F=2048$ dimensional face descriptor,
while we also experiment with reducing the dimensionality
via an additional FC, down to $D_F=256$ or $D_F=128$.

To disambiguate various network configurations,
we name the networks as \emph{Ext}-GV-$S$(-g$G$),
where \emph{Ext} is the feature extractor network
(\emph{Res} for ResNet-50 or \emph{SE} for SENet-50),
$S$ is the size of image sets used during training,
and $G$ is the number of ghost clusters (if zero,
the suffix is dropped).
For example, \emph{SE-GV-3-g2} denotes a network which uses
the SENet-50 as the feature extractor,
training image sets of size 3, and
2 ghost clusters.

\paragraph{Network deployment.}
In the IJB-A and IJB-B datasets, there are images and
videos available for each subject.
Here we follow the established approach of~\cite{Cao18,Crosswhite17}
to balance the contributions of face examples from
different sources, as otherwise a single very long video could
completely dominate the representation.
In more detail, face examples are extracted from
all video frames, and their additive contributions
to the GhostVLAD representation are down-weighed by
the number of frames in the video.

The similarity between two templates is measured as the
scalar product between the template representations;
recall that they have unit norm (Fig.~\ref{fig:cnn}).

\paragraph{Baselines.}
Our network is compared with several average-pooling baselines.
The baseline architecture consists of a feature extractor network
which produces a face descriptor for each input example,
and the template representation is performed by
average-pooling the face descriptors
(with source balancing), followed by
L2 normalization.
The same feature extractor networks are used
as for our method, ResNet-50 or SENet-50,
abbreviated as \emph{Res} and \emph{SE}, respectively,
with an optionally added FC layer to perform dimensionality reduction
down to 128-D or 256-D.
These networks are trained for single-image face classification,
which is equivalent to stage 1
of our training procedure from Sec.~\ref{sec:imp},
and also corresponds to the current state-of-the-art
approach~\cite{Cao18} (albeit with more training data
-- see Sec.~\ref{sec:exp:sota} for details and comparisons).
No image degradation is performed as it decreases performance
when combined with single-image classification training.

In addition, we train the baseline architecture SENet-50 with
average-pooling
using our training procedure (Sec.~\ref{sec:train}),
\ie with image sets of size 2 and degraded images,
and refer to it as \emph{SE-2}.

\subsection{Ablation studies on IJB-B}

Here we evaluate various design choices of our architecture
and compare it to baselines on the IJB-B dataset,
as it is larger and more challenging than IJB-A;
results on verification and identification are shown in
Tables~\ref{tab:ijbb-ver} and Table~\ref{tab:ijbb-id},
respectively.

\paragraph{Feature extractor and dimensionality reduction.}
Comparing rows 1 \vs 2 of the two tables shows
that reducing the dimensionality of the face features
from 2048-D to 128-D does not affect the performance much,
and in fact sometimes improves it due to added parameters
in the form of the dimensionality reduction FC.
As the feature extractor backbone,
SENet-50 consistently beats ResNet-50, 
as summarized in rows 2 \vs 3.

\newcounter{magicrownumbers} 
\newcommand\rownumber{\stepcounter{magicrownumbers}
\arabic{magicrownumbers}}

\begin{table*}[t!]
\captionsetup{font=small}
\begin{center}{\scalebox{0.80}{
\setlength{\tabcolsep}{6pt}
\begin{tabular}{c|c|c|c|c|c|c|c|c|c|c|c}
\hline
Row & Network & $D_F$ & $D$ & $K$ & $G$ & No.\ & Deg.\ & \multicolumn{4}{c}{1:1 Verification TAR (FAR=)} \\
id & & & & & & faces &  & $1E-5$ & $1E-4$ & $1E-3$ &$1E-2$ \\
\hline

\rownumber & Res~\cite{Cao18} & 2014 & 2048 & - & - & 1 & \xmark & $0.647$ & $0.784$ & $0.878$ &  $0.938$ \\

\rownumber & Res & 128 & 128 & - & - & 1 & \xmark & 0.646 & 0.785 & 0.890 & 0.954 \\
\rownumber & SE & 128 & 128 & - & - & 1 & \xmark & 0.670 & 0.803 & 0.896 & 0.954 \\
\rownumber & SE & 256 & 256 & - & - & 1 & \xmark & 0.677 & 0.807 & 0.892 & 0.955 \\ 
\rownumber & SE-2 & 256 & 256 & - & - & 2 & \cmark & 0.679 & 0.810 & 0.902 & 0.958 \\ \hline 

\rownumber  & Res-GV-2 & 128 & 128 & 8 & 0 & 2 & \cmark & 0.715 & 0.835 & 0.916 & 0.963 \\
\rownumber & SE-GV-2 & 128 & 128 & 8 & 0 & 2 & \cmark & 0.721 & 0.835 & 0.916 & 0.963 \\ 
\rownumber & SE-GV-2 & 256 & 128 & 8 & 0 & 2 & \xmark & 0.685 & 0.823 & 0.925 & 0.963 \\ 
\rownumber & SE-GV-2 & 256 & 128 & 8 & 0 & 2 & \cmark & 0.738 & 0.850 & 0.923 & 0.964 \\ 
\rownumber & SE-GV-2 & 256 & 128 & 4 & 0 & 2 & \cmark & 0.729 & 0.841 & 0.914 & 0.957 \\ 
\rownumber & SE-GV-2 & 256 & 128 & 16 & 0 & 2 & \cmark & 0.722 & 0.848 & 0.921 & 0.964 \\ \hline 

\rownumber & SE-GV-3 & 256 & 128 & 8 & 0 & 3 & \cmark & 0.741 & 0.853 & 0.925 & 0.963 \\
\rownumber & SE-GV-4 & 256 & 128 & 8 & 0 & 4 & \cmark & 0.747 & 0.852 & 0.922 & 0.961 \\ 
\rownumber & SE-GV-3-g1 & 256 & 128 & 8 & 1 & 3 & \cmark & 0.753 & 0.861 & $\mathbf{0.926}$ & 0.963 \\ 
\rownumber & SE-GV-4-g1 & 256 & 128 & 8 & 1 & 4 & \cmark & $\mathbf{0.762}$ & $\mathbf{0.863}$ & $\mathbf{0.926}$ & 0.963 \\ 
\rownumber & SE-GV-3-g2 & 256 & 128 & 8 & 2 & 3 & \cmark & 0.754 & 0.861 & 0.926 & $\mathbf{0.964}$  \\ 

\rownumber & SE-GV-4 & 256 & 256 & 8 & 0 & 4 & \cmark & 0.713 & 0.838 & 0.919 & 0.963 \\ 
\rownumber & SE-GV-4-g1 & 256 & 256 & 8 & 1 & 4 & \cmark & 0.739 & 0.853 & 0.924 & 0.963 \\

\hline
\end{tabular}}}
\end{center}
\vspace{-1.5mm}
\caption{
	\textbf{Verification performance
on the IJB-B dataset.} 
A higher value of TAR is better.
$D_F$ is the face descriptor dimension before aggregation.
$D$ is the dimensionality of the final template representation.
$K$ and $G$ are the number of non-ghost and ghost clusters in GhostVLAD,
respectively.
`No.\ faces' is the number of 
faces per set used during training. `Deg.' indicates whether 
the training images are degraded.
All training is done using the VGGFace2 dataset.
}
\label{tab:ijbb-ver}
\vspace{-5mm}
\end{table*}

\begin{table*}[t!]
\captionsetup{font=small}
\begin{center}{\scalebox{0.80}{
\setlength{\tabcolsep}{4pt}
\begin{tabular}{c|c|c|c|c|c|c|c|c|c|c|c|c}
\hline
Row & Network & $D_F$ & $D$ & $K$ & $G$ & No.\ & Deg.\ & \multicolumn{5}{c}{1:N Identification TPIR}\\
id &	 & & & & & faces & & FPIR= & FPIR= & & & \\
   & & & & & & & & $0.01$ & $0.1$ & Rank-$1$ & Rank-$5$& Rank-$10$ \\
\hline

\rownumber & Res~\cite{Cao18} & 2048 & 2048 & - & - & 1 & \xmark & $0.701$& $0.824$ & $0.886$ & $0.936$ & $0.953$\\

\rownumber & Res & 128 & 128 & - & - & 1 & \xmark & 0.688 & 0.833 & 0.901 & 0.950 & 0.963 \\
\rownumber & SE & 128 & 128 & - & - & 1 & \xmark & 0.712 & 0.849 & 0.908 & 0.949 & 0.963 \\
\rownumber & SE & 256 & 256 & - & - & 1 & \xmark & 0.718 & 0.854 & 0.908 & 0.948 & 0.962 \\ 
\rownumber & SE-2 & 256 & 256 & - & - & 2 & \cmark & 0.717 & 0.857 & 0.909 & 0.949 & 0.962 \\  \hline

\rownumber & Res-GV-2 & 128 & 128 & 8 & 0 & 2 & \cmark & 0.762 & 0.872 & 0.917 & 0.953 & 0.964 \\
\rownumber & SE-GV-2 & 128 & 128 & 8 & 0 & 2 & \cmark & 0.753 & 0.880 & 0.917 & 0.953 & 0.964 \\ 
\rownumber & SE-GV-2 & 256 & 128 & 8 & 0 & 2 & \xmark & 0.751 & 0.884 & 0.912 & 0.952 & 0.962 \\
\rownumber & SE-GV-2 & 256 & 128 & 8 & 0 & 2 & \cmark & 0.760 & 0.879 & 0.918 & 0.955 & 0.964\\ 
\rownumber & SE-GV-2 & 256 & 128 & 4 & 0 & 2 & \cmark & 0.749 & 0.868 & 0.914 & 0.953 & 0.963 \\ 
\rownumber & SE-GV-2 & 256 & 128 & 16 & 0 & 2 & \cmark & 0.759 & 0.879 & 0.918 & 0.954 & $\mathbf{0.965}$ \\ \hline

\rownumber & SE-GV-3 & 256 & 128 & 8 & 0 & 3 & \cmark & 0.764 & 0.885 & 0.921 & 0.955 & 0.962 \\
\rownumber & SE-GV-4 & 256 & 128 & 8 & 0 & 4 & \cmark & 0.752 & 0.878 & 0.914 & 0.952 & 0.960 \\ 
\rownumber & SE-GV-3-g1 & 256 & 128 & 8 & 1 & 3 & \cmark & 0.770 & $\mathbf{0.888}$ & $\mathbf{0.923}$ & $0.956$ & $\mathbf{0.965}$ \\ 
\rownumber & SE-GV-4-g1 & 256 & 128 & 8 & 1 & 4 & \cmark & $\mathbf{0.776}$ & $\mathbf{0.888}$ & 0.921 & $\mathbf{0.957}$ & 0.964 \\
\rownumber & SE-GV-3-g2 & 256 & 128 & 8 & 2 & 3 & \cmark & 0.772 & 0.886 & 0.922 & $\mathbf{0.957}$ & 0.964 \\ 
\rownumber & SE-GV-4 & 256 & 256 & 8 & 0 & 4 & \cmark & 0.732 & 0.870 & 0.912 & 0.952 & 0.963 \\ 
\rownumber & SE-GV-4-g1 & 256 & 256 & 8 & 1 & 4 & \cmark & $\mathbf{0.776}$ & 0.883 & 0.921 & $\mathbf{0.957}$ & $\mathbf{0.965}$ \\

\hline
\end{tabular}}}
\end{center}
\vspace{-1.5mm}
\caption{\textbf{Identification performance on the IJB-B dataset.} 
A higher value of TPIR is better.
See caption of Tab.~\ref{tab:ijbb-ver} for the explanations of column titles.
Note, for readability  standard 
deviations are not included here, but are included in Tab.~\ref{tab:ijba-id}.
}
\label{tab:ijbb-id}
\vspace{-1cm}
\end{table*}

\paragraph{Training for set-based face recognition.}
The currently adopted set-based face recognition approach
of training with single-image examples
and performing aggregation post hoc (\emph{SE}, row 4)
is clearly inferior to
our training procedure which is aware of image sets
(\emph{SE-2}, row 5).

\paragraph{Learnt GhostVLAD aggregation.}
Using the GhostVLAD aggregation layer
(with $G=0$ \ie equivalent to NetVLAD)
together with our set-based training framework strongly outperforms
the standard average-pooling approach,
regardless of whether training is done with non-degraded images
(\emph{SE-GV-2}, row 8 \vs \emph{SE}, rows 3 and 4),
degraded images
(\emph{SE-GV-2}, row 9 \vs \emph{SE-2}, row 5),
or if a different feature extractor architecture (ResNet-50)
is used (\emph{Res-GV-2}, row 6 \vs \emph{Res}, row 2).
Using 256-D \vs 128-D face descriptors as inputs to GhostVLAD,
while keeping the same dimensionality of the final template representation
(128-D),
achieves better results (rows 9 \vs 7),
so we use 256-D in all latter experiments.

\paragraph{Training with degraded images.}
When using our set-based training procedure,
training with degraded images brings a consistent boost,
as shown in rows 9 \vs 8, since it better matches the test-time
scenario which contains images of varying quality.

\paragraph{Number of clusters $K$.}
GhostVLAD (and NetVLAD) have a hyperparameter $K$
-- the number of non-ghost clusters -- which we vary
between 4 and 16 (rows 9 to 11) to study its effect on
face recognition performance.
It is expected that $K$ shouldn't be too small
so that underfitting is avoided
(\eg $K=1$ is similar to average-pooling)
nor too large in order to prevent over-quantization and overfitting.
As in traditional image retrieval~\cite{Jegou10},
we find that a wide range of $K$ achieves good performance,
with $K=8$ being the best.

\paragraph{Ghost clusters.}
Introducing a single ghost cluster ($G=1$) brings significant improvement
over the vanilla NetVLAD, as shown by comparing
\emph{SE-GV-3-g1} \vs \emph{SE-GV-3} (rows 14 \vs 12)
and
\emph{SE-GV-4-g1} \vs \emph{SE-GV-4} (rows 15 \vs 13).
Using one ghost cluster is sufficient as
increasing the number of ghost clusters to two does not result in
significant differences (row 16 \vs row 14).
Ghost clusters enable the system to automatically down-weight
the contribution of low quality images, as will be shown in
Sec.~\ref{sec:visualize_ghost}, which improves the template representations
and benefits face recognition.

\paragraph{Set size used during training.}
To perform set-based training, as described in Sec.~\ref{sec:imp},
image sets are created by sampling a fixed number of faces for a subject;
the number of sampled faces is another parameter of the method.
Increasing the set size from 2 to 3 consistently improves results
(rows 9 \vs 12), while there is no clear winner between using
3 or 4 face examples (worse for $G=0$, rows 12 \vs 13,
better for $G=1$, rows 15 \vs 14).

\paragraph{Output dimensionality.}
Comparisons are also made between networks 
with 128-D output features and those with 256-D 
(\ie row 13 \vs 17 and row 15 \vs 18),
and we can see that networks with 128-D output achieve 
better performance while being more memory-efficient.

\subsection{Comparison with state-of-the-art}\label{sec:exp:sota}
\begin{table*}[t]
\captionsetup{font=small}
\begin{center}{\scalebox{0.80}{
\begin{tabular}{c|c|c|c|c|c|c|c}
\hline
Network & Training & $D$ & \multicolumn{5}{c}{1:1 Verification TAR} \\
	       &	dataset	  & & FAR=$1E-5$ &  FAR=$1E-4$ & FAR=$1E-3$ & FAR=$1E-2$ & FAR=$1E-1$\\
\hline
\multicolumn{8}{c}{IJB-A} \\ \hline

Bin~\cite{Hassner16} & ImNet+CAS & 4096 & - & - & 0.631 & 0.819 & - \\
NAN~\cite{Yang17} & priv1 & 128 & - & - & $0.881 \pm 0.011$ & $0.941 \pm 0.008$ & $0.978 \pm 0.003$ \\
QAN~\cite{Liu17} & VF+priv2 & - & - & - & $0.893 \pm 0.039$ & $0.942 \pm 0.015$ & $0.980 \pm 0.006$ \\

DREAM~\cite{Cao18a} & MS & - & - & - & $0.868 \pm 0.015$  & $0.944 \pm 0.009$ & - \\
SF+R~\cite{Zheng18} & MS-clean & 512-pi & - & - & $0.932$ & - & - \\
MN-vc~\cite{Xie18b} & VF2 & 2048 & - & - & $0.920 \pm 0.013$  & $0.962 \pm 0.005$ & $0.989 \pm 0.002$ \\

SE~\cite{Cao18} & VF2 & 2048 & - & - & $0.904 \pm 0.020$ & $0.958 \pm 0.004$ & $0.985 \pm 0.002$ \\
SE~\cite{Cao18} & MS+VF2 & 2048  & - & - & $0.921\pm 0.014$ & $0.968 \pm 0.006$ & $0.990 \pm 0.002$ \\

SE-GV-3 & VF2 & 128 & - & - & $\mathbf{0.935 \pm 0.016}$ & $\mathbf{0.972 \pm 0.005}$ & $0.988 \pm 0.002$ \\
SE-GV-4-g1 & VF2 & 128 & - & - & $\mathbf{0.935 \pm 0.015}$ & $\mathbf{0.972 \pm 0.003}$ & $\mathbf{0.990 \pm 0.002}$ \\ \hline

\multicolumn{8}{c}{IJB-B} \\ \hline
SE~\cite{Cao18} & VF2 & 2048 &  $0.671$ & $0.800$ & $0.888$ & $0.949$ &  -\\
SE~\cite{Cao18}  & MS+VF2 & 2048 & 0.705 & 0.831 & 0.908 &0.956 & -\\

MN-vc~\cite{Xie18b} & VF2 & 2048 & 0.708 & 0.831 & $0.909$  & $0.958$ & - \\
SE+DCN~\cite{Xie18a} & VF2 & - & 0.730* & 0.849 & $\mathbf{0.937}$  & $\mathbf{0.975}$ & - \\

SE-GV-3 & VF2 & 128 & 0.741 & 0.853 & 0.925 & 0.963 & - \\
SE-GV-4-g1 & VF2  & 128 & $\mathbf{0.762}$ & $\mathbf{0.863}$ & $0.926$ & 0.963 & -\\

\hline
\end{tabular}}}
\end{center}
\vspace{-1.5mm}
\caption{\textbf{Comparison with state-of-the-art for
{\em verification} on the IJB-A and IJB-B datasets.}
A higher value of TAR is better.
$D$ is the dimension of the 
template representation.
The training datasets abbreviations are VGGFace~\cite{Parkhi15} (VF), 
VGGFace2~\cite{Cao18} (VF2), MS-Celeb-1M~\cite{Guo16} (MS),
a cleaned subset of MS-Celeb-1M refined by~\cite{Zheng18} (MS-clean),
ImageNet~\cite{Russakovsky15} and CASIA WebFace~\cite{Yi14} (ImNet+CAS),
and private datasets used by~\cite{Yang17} (priv1) and~\cite{Liu17} (priv2).
`512-pi' means that a 512-D descriptor 
is used per image. `*' denotes the value given by the author.
Our best network, SE-GV-4-g1, sets the state-of-the-art by a significant
margin on both datasets (except for concurrent work~\cite{Xie18a}).
}
\label{tab:ijba-ver}
\vspace{-5mm}
\end{table*}

\begin{table*}[t]
\captionsetup{font=small}
\begin{center}{\scalebox{0.80}{
\begin{tabular}{c|c|c|c|c|c|c|c}
\hline
Network & Training & $D$ & \multicolumn{5}{c}{1:N Identification TPIR}\\
	& dataset	  & & FPIR=$0.01$ & FPIR=$0.1$ & Rank-$1$ & Rank-$5$& Rank-$10$\\
\hline
\multicolumn{8}{c}{IJB-A} \\ \hline

Bin~\cite{Hassner16} & ImNet+CAS & 4096 & 0.875 & - & 0.846 & 0.933 & 0.951 \\
NAN~\cite{Yang17} & priv1 & 128 &  $0.817 \pm 0.041$ & $0.917 \pm 0.009$ & $0.958 \pm 0.005$ & $0.980 \pm 0.005$ & $0.986 \pm 0.003$ \\

DREAM~\cite{Cao18a} & MS & - & - & - & $0.946 \pm 0.011$  & $0.968 \pm 0.010$ & - \\

SE~\cite{Cao18} & VF2 & 2048 & $0.847 \pm 0.051$ & $0.930 \pm 0.007$ & $0.981\pm 0.003$ & $\mathbf{0.994 \pm 0.002}$  & $\mathbf{0.996 \pm 0.001}$ \\
SE~\cite{Cao18} & MS+VF2 & 2048  & $0.883 \pm 0.038$ & $0.946 \pm 0.004$ & $\mathbf{0.982 \pm 0.004}$ & $0.993 \pm 0.002$ & $0.994 \pm 0.001$\\

SE-GV-3 & VF2 & 128  & $0.872 \pm 0.066$ & $0.951 \pm 0.007$  & $0.979 \pm 0.005$ & $0.990 \pm 0.003$ & $0.992 \pm 0.003$ \\
SE-GV-4-g1 & VF2 & 128  & $\mathbf{0.884 \pm 0.059}$ & $\mathbf{0.951 \pm 0.005}$ & $0.977 \pm 0.004$ & $0.991 \pm 0.003$  & $0.994 \pm 0.002$ \\ \hline

\multicolumn{8}{c}{IJB-B} \\ \hline
SE~\cite{Cao18}  & VF2 &  2048 &   $0.706 \pm 0.047 $ & $0.839 \pm 0.035$ & $0.901 \pm 0.030$ & $0.945 \pm 0.016$ & $0.958 \pm 0.010$ \\ 
SE~\cite{Cao18}  & MS+VF2 &  2048  & $0.743 \pm 0.037$ & $0.863 \pm 0.032$ & $0.902 \pm 0.036$ & $0.946 \pm 0.022$ & $0.959 \pm 0.015$ \\

SE-GV-3 & VF2 & 128  & $0.764 \pm 0.041$ & $0.885 \pm 0.032$ & $0.921 \pm 0.023$ & $0.955 \pm 0.013$ & $0.962 \pm 0.010$ \\
SE-GV-4-g1 & VF2 & 128  & $\mathbf{0.776 \pm 0.030}$ & $\mathbf{0.888 \pm 0.029}$ & $\mathbf{0.921 \pm 0.020}$ & $\mathbf{0.956 \pm 0.013}$ & $\mathbf{0.964 \pm 0.010}$ \\

\hline
\end{tabular}}}
\end{center}
\vspace{-1.5mm}
\caption{\textbf{Comparison with state-of-the-art for
{\em identification} on the IJB-A and IJB-B datasets.}
A higher value of TPIR is better. 
See caption of Tab.~\ref{tab:ijba-ver} for explanations of abbreviations.
Our best network, SE-GV-4-g1, sets the state-of-the-art by a significant
margin on both datasets.
}
\label{tab:ijba-id}
\vspace{-4mm}
\end{table*}

In this section, our best networks,
\emph{SE-GV-3} and \emph{SE-GV-4-g1},
are compared against the state-of-the-art
on the IJB-A and IJB-B datasets.
The currently best performing method~\cite{Cao18}
is the same as our \emph{SE} baseline
(\ie average-pooling of SENet-50 features
trained for single-image classification)
but trained on a much larger training set,
MS-Celeb-1M dataset~\cite{Guo16},
and then fine-tuned on VGGFace2.

\begin{figure*}[t]
\captionsetup{font=small}
\begin{center}
\begin{tabular}{c@{~~~~~}c}
Verification ROC & Identification DET \\
{\scriptsize (higher is better)} & {\scriptsize (lower is better)} \\ [0.1cm]
\includegraphics[align=c,width=44mm,clip,trim=0 0 0 1cm]{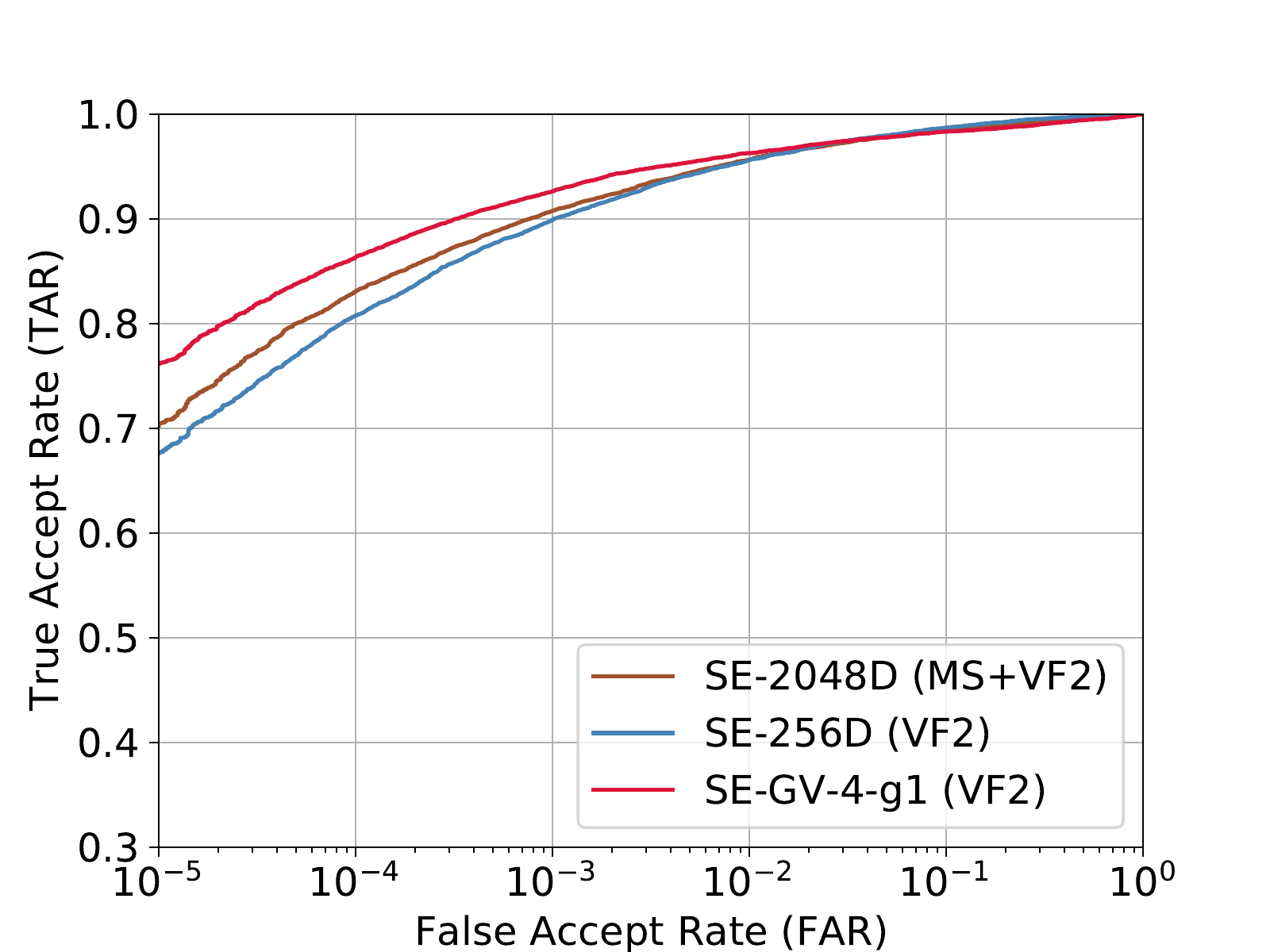} &
\includegraphics[align=c,width=42mm,clip,trim=0 0.5cm 0 0]{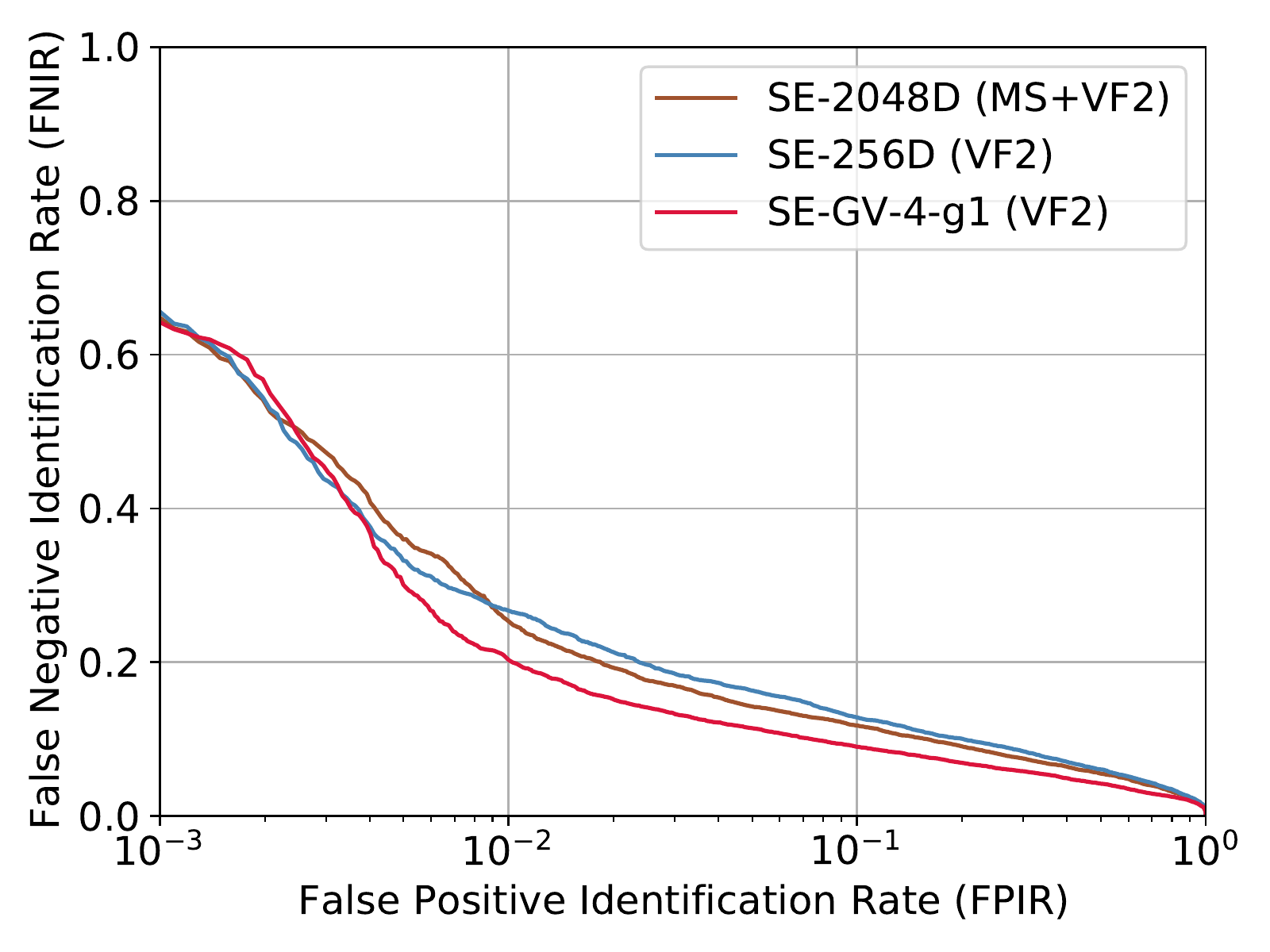}
\end{tabular}
\end{center}
\vspace{-2mm}
\caption{{\bf Results on the IJB-B dataset.} 
Our \emph{SE-GV-4-g1} network which produces 128-D templates,
beats the best baseline (\emph{SE} with 256-D templates)}
and the state-of-the-art trained on a much larger dataset
(\emph{SE} with 2048-D templates) trained on VGGFace2 and MS-Celeb-1M).
\label{fig:ijb}
\vspace{-4mm}
\end{figure*}

From Tables~\ref{tab:ijba-ver} and \ref{tab:ijba-id},
and Figure~\ref{fig:ijb},
it is clear our GhostVLAD network (\emph{SE-GV-4-g1})
convincingly outperforms previous methods and sets
the new state-of-the-art for both identification and
verification on both IJB-A and IJB-B datasets.
In particular, it surpasses~\cite{Zheng18} marginally
on the IJB-A verification task,  
despite the fact that~\cite{Zheng18}
uses a deeper ResNet and performs an exhaustive
scoring using each face image in the templates.
The only points for which the GhostVLAD network 
doesn't beat the state-of-the-art, 
though it is on par with it, 
is in TPIR at Rank-1 to Rank-10
for identification on IJB-A;  but this is because
IJB-A is not challenging enough and the TPIR values have
saturated to a 99\% mark.
For the same measures on the more challenging IJB-B benchmark, 
our network achieves the best TAR at FAR=$1E-5$ and FAR=$1E-4$,
and is only lower than a concurrent work~\cite{Xie18a} at FAR=$1E-3$ and FAR=$1E-2$.
Furthermore, our networks produce much 
smaller template descriptors than the previous state-of-the-art 
networks (128-D \vs 2048-D),
making them more useful 
in real-world applications due to smaller memory requirements
and faster template comparisons.

The results are especially impressive as we only train
using VGGFace2~\cite{Cao18} and beat methods which train with much more
data, such as~\cite{Cao18}
which combine VGGFace2 and MS-Celeb-1M~\cite{Guo16},
\eg TAR at FAR=$1E-5$ of $0.762$ \vs $0.705$ for verification on IJB-B,
and TPIR at FPIR=0.01 of $0.776$ \vs $0.743$ for identification on IJB-B.
When considering only methods trained on the same data (VGGFace2),
our improvement over the state-of-the-art is even larger:
TAR at FAR=$1E-5$ of $0.762$ \vs $0.671$ for verification on IJB-B,
and TPIR at FPIR=0.01 of $0.776$ \vs $0.706$ for verification on IJB-B.

\subsection{Analysis of ghost clusters} \label{sec:visualize_ghost}

\begin{figure}[t]
   \begin{center}
         \includegraphics[width=0.90\columnwidth]{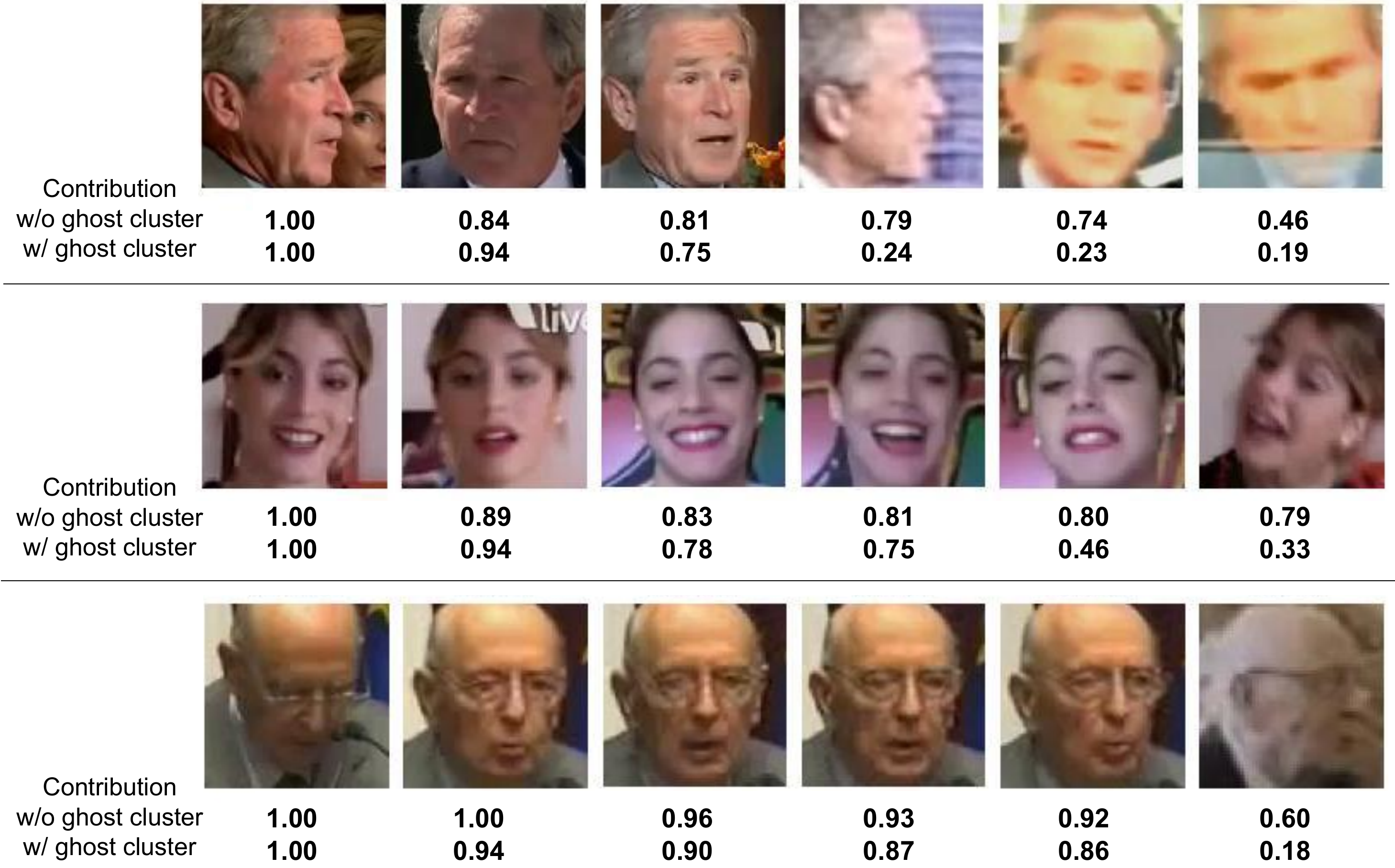}
   \end{center}
   \caption{\textbf{Effect of ghost clusters.}
Each row shows shows 6 images from a template in the IJB-B dataset.
The contribution (relative to the max) of each image
to the final template representation is shown
(see Sec.~\ref{sec:visualize_ghost} for details),
for the cases of no ghost clusters ($G=0$, network \emph{SE-GV-3})
and one ghost cluster ($G=1$, network \emph{SE-GV-4-g1})
in GhostVLAD.
Introduction of a single ghost cluster dramatically reduces
the contribution of low-quality images to the template,
improving the signal-to-noise ratio.
}
    \label{fig:score}
\end{figure}

Addition of ghost clusters was motivated by the intuition
that it enables our network to learn to ignore uninformative
low-quality images by assigning them to the discarded ghost clusters.
Here we evaluate this hypothesis qualitatively.

Recall that GhostVLAD computes a template representation
by aggregating residual vectors of input descriptors,
where a residual vector is a concatenation of
per non-ghost cluster residuals
weighted by their non-ghost assignment weights (Sec.~\ref{sec:ghost}).
Therefore, the contribution of a specific example image towards
the template representation can be measured as the norm of the residual.

Figure~\ref{fig:score} show that our intuition is correct --
the network automatically learns to dramatically down-weight
blurry and low-resolution images, thus improving the signal-to-noise ratio.
Note that this behaviour emerges completely automatically
without ever explicitly teaching the network to down-weight low-quality
images.

\section{Conclusions}\label{sec:conclusion}
We introduced a neural network
architecture and training procedure
for learning compact representations of image sets
for template-based face recognition.
Due to the novel GhostVLAD layer, the network is able to
automatically learn to weight face descriptors
depending on their information content.
Our template representations outperform
the state-of-the-art on the challenging IJB-A and IJB-B benchmarks
by a large margin. 

The network architecture proposed here could also be applied to other
image-set tasks such as person re-identification, and set-based
retrieval. More generally, the idea of having a `null' vector available for
assignments could have applicability in many situations where it is advantageous to have 
a mechanism to remove noisy or corrupted data.

\paragraph{Acknowledgements}
We thank Weidi Xie for his useful advice, 
and we thank Li Shen for providing pre-trained
networks.
This work was funded by an EPSRC studentship and 
EPSRC Programme Grant Seebibyte EP/M013774/1.

\bibliographystyle{splncs04}
\bibliography{shortstrings,mybib,vgg_local,vgg_other}

\end{document}